\RequirePackage{amsmath}
\documentclass[a4paper,oribibl]{llncs}

\usepackage{amssymb,amsmath,amstext}
\usepackage{algorithm2e}
\usepackage[noend]{algorithmic}
\usepackage{array}
\usepackage{graphicx}
\usepackage{multirow}
\usepackage{tensor}
\usepackage{url}
\usepackage{graphicx}
\usepackage{multirow}
\usepackage[dvipsnames]{xcolor}
\usepackage{ulem}
\usepackage{rotating}
\usepackage{fixmath}
\usepackage{lipsum}
\usepackage{enumerate}

\graphicspath{{images/}}

\newcommand{\keywords}[1]{\par\addvspace\baselineskip\noindent\keywordname\enspace\ignorespaces#1}

\begin{document}

\mainmatter

\title{Markerless Inside-Out Tracking for Interventional Applications}

\author{
Benjamin Busam${}^{1,2}$,
Patrick Ruhkamp${}^{1,2}$,
Salvatore Virga${}^{1}$,\\
Beatrice Lentes${}^{1}$,
Julia Rackerseder${}^{1}$,\\
Nassir Navab${}^{1,3}$,
Christoph Hennersperger${}^{1}$
}

\institute{
${}^{1}$ Computer Aided Medical Procedures, Technische Universit\"at M\"unchen, Germany\\
${}^{2}$ FRAMOS GmbH, Germany\\
${}^{3}$ Computer Aided Medical Procedures, Johns Hopkins University, US\vspace{2ex}\\
{\tt\small b.busam@framos.com},
{\tt\small p.ruhkamp@framos.com},
{\tt\small salvo.virga@tum.de},\\
{\tt\small beatrice.lentes@tum.de},
{\tt\small julia.rackerseder@tum.de},\\
{\tt\small navab@cs.tum.edu},
{\tt\small christoph.hennersperger@tum.de}
}

\maketitle

\begin{abstract}
Tracking of rotation and translation of medical instruments plays a substantial role in many modern interventions. Traditional external optical tracking systems are often subject to line-of-sight issues, in particular when the region of interest is difficult to access or the procedure allows only for limited rigid body markers. The introduction of inside-out tracking systems aims to overcome these issues. We propose a marker-less tracking system based on visual SLAM to enable tracking of instruments in an interventional scenario. To achieve this goal, we mount a miniature multi-modal (monocular, stereo, active depth) vision system on the object of interest and relocalize its pose within an adaptive map of the operating room. We compare  state-of-the-art algorithmic pipelines and apply the idea to transrectal 3D Ultrasound (TRUS) compounding of the prostate. Obtained volumes are compared to reconstruction using a commercial optical tracking system as well as a robotic manipulator. Feature-based binocular SLAM is identified as the most promising method and is tested extensively in challenging clinical environment under severe occlusion and for the use case of prostate US biopsies.
\end{abstract}


\section{Introduction} 
\label{sec:intro}
Tracking of medical instruments and tools is required for various systems in medical imaging, as well as computer aided interventions.
In general, tracking systems provide the rigid body transformation of one (or multiple) targets with respect to a common reference frame, which can be the patient, a camera system, or any pre-calibrated coordinate space.
Especially for medical applications, accurate tracking is an important requirement, however often comes with severe drawbacks impacting the medical workflow.
Mechanical tracking systems (robotic arms or linear stages) can provide highly precise tracking through a kinematic chain~\cite{hennersperger2017towards},~\cite{Adebar2011}.
However, these systems often require bulky and expensive equipment, which cannot be adapted to a clinical environment where high flexibility needs to be ensured. 
In contrast to that, electromagnetic tracking is flexible in its use, but is limited to comparably small work spaces and can interfere with metallic objects in proximity to the target, severely reducing the accuracy~\cite{kral2013comparison}.

\begin{figure}[t]
\centering
\includegraphics[width=0.8\textwidth]{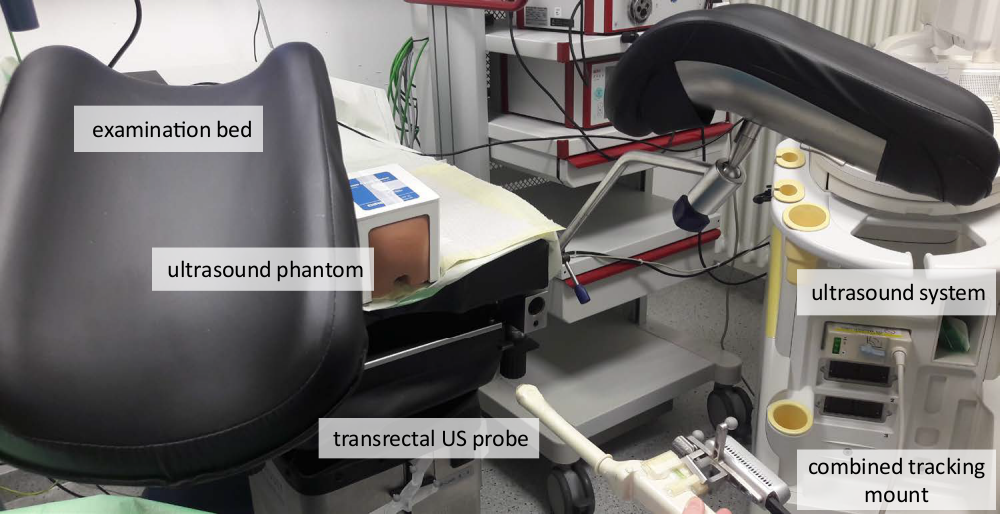}
	\caption{\textbf{Interventional setup for fusion biopsy.} Clinical settings are often characterized by cluttered setups with tools and equipment around the examination bed. While such environments are challenging for outside-in tracking, they can provide a rich set of features for SLAM-based inside-out tracking.}
	\label{fig:interventionalSetup}
\end{figure}

Optical tracking systems (OTS) enjoy widespread use as they do not have these disadvantages. Usually, a set of active or passive infrared markers is attached to the target and tracked by a static external stereo camera.
Despite favourable spatial accuracy under optimal conditions, respective systems suffer from constraints by the required line-of-sight.
Especially in interventional settings, this can impair the positioning of equipment and staff, as occlusions of the markers occur while they are in use.
Robust marker based methods such as~\cite{busam2015stereo} address this problem and work even if the target is only partly visible. However, the marker-visibility issue is further complicated for imaging solutions relying on tracking systems, with prominent examples being freehand SPECT imaging~\cite{heuveling2012sentinel} as well as freehand 3D ultrasound imaging~\cite{fenster2001three}.

Aiming at both accurate and flexible systems for 3D imaging, a series of developments have been proposed recently.
Inside-out tracking for collaborative robotic imaging~\cite{esposito2015cooperative} proposes a marker-based approach using infrared cameras, however, not  resolving line-of-sight issues.
A first attempt at making use of localized features employs tracking of specific skin features for estimation of 3D poses~\cite{sun2014probe} in 3D US imaging.
While this work shows promising results, it is constrained to the specific anatomy at hand.

In contrast to previous works, our aim is to provide a generalizable tracking approach without requiring a predefined or application-specific set of features.
With the recent advent of advanced miniaturized camera systems, our aim is to evaluate an inside-out tracking approach solely relying on features extracted from image sensor data for pose tracking.
%
For a generic inside-out tracking approach, which is robust in different environments, the sole geometric information of the scenery without any prior knowledge, is vital for orientation.
For this purpose, we propose the use of visual methods for simultaneously mapping the scenery and localizing the system within it. This is enabled by building up a map from characteristic structures within the previously unknown scene observed by a camera, which is known as SLAM~\cite{ORB-SLAM2_Mur-Artal_TOR2017}.
SLAM methods can be distinguished between direct and feature-based methods, both with its characteristic drawbacks and benefits. 
For direct SLAM approaches, the whole image information is taken into account~\cite{LSD-SLAM_Engel_ECCV2014,DSO_Engel_PAMI2018}, which may lead to erroneous poses under changing lighting conditions, require good initialization and are not able to recover poses correctly for rolling shutter cameras.
In contrast, feature-based methods rely on extracted feature points~\cite{ORB-SLAM2_Mur-Artal_TOR2017}, which lead to more stable tracking behaviour during illumination changes, but require a minimum amount of structure within the scene. 
Different image modalities
can be used for visual SLAM, whereas a stereo setup possesses many benefits compared to monocular vision or active depth sensors.

\begin{figure}[t]
\centering
\includegraphics[width=\textwidth]{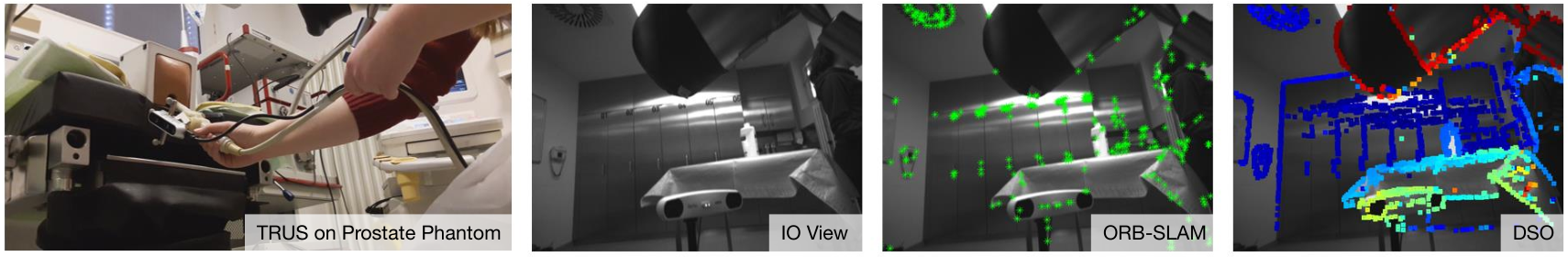}
	\caption{\textbf{3D TRUS volume acquisition of prostate phantom.} The inside-out camera is mounted on the transrectal US transducer together with a rigid marker for an outside-in system in the prostate biopsy OR. The consecutive images show the relevant extracted data for the considered SLAM methods.}
	\label{fig:prostateRoom}
\end{figure}

On this foundation, we propose a flexible inside-out tracking approach relying on image features and poses retrieved from SLAM.
We evaluate different methods in direct comparison to a commercial tracking solution and ground truth, and show an integration for freehand 3D US imaging as one potential use-case.
The proposed prototype is the first proof of concept for SLAM-based inside-out tracking for interventional applications, applied here to 3D TRUS as shown in Fig.~\ref{fig:prostateRoom}. The novelty of pointing the camera away from the patient into the quasi-static room while constantly updating the OR map enables advantages in terms of robustness, rotational accuracy and line-of-sight problem avoidance. Thus, no hardware relocalization of external outside-in systems is needed, partial occlusion is handled with wide-angle lenses and the method copes with dynamic environmental changes. Moreover, it paves the path for automatic multi-sensor alignment through a shared common map while maintaining an easy installation by clipping the sensor to tools.

\section{Methods}
\label{sec:methods}
For interventional imaging and specifically for the case of 3D ultrasound, the goal is to provide rigid body transformations of a desired target with respect to a common reference frame.
This way, we denote $\tensor[^{B}]{\textbf{T}}{_{A}}$ as transformation $A$ to $B$.
On this foundation, the transformation $\tensor[^{W}]{\textbf{T}}{_{US}}$ from the ultrasound image ($US$) should be indicated in a desired world coordinate frame ($W$).
For the case of inside-out based tracking - and in contrast to outside-in approaches - the ultrasound probe is rigidly attached to the camera system, providing the desired relation to the world reference frame 
\begin{equation}
    \tensor[^{W}]{\textbf{T}}{_{US}} = 
    \tensor[^{W}]{\textbf{T}}{_{RGB}} \cdot
    \tensor[^{RGB}]{\textbf{T}}{_{US}},
\end{equation}
where $\tensor[^{W}]{\textbf{T}}{_{RGB}}$ is retrieved from tracking (see Sec. \ref{sec:insideOutTracking}). The static transformation $\tensor[^{RGB}]{\textbf{T}}{_{US}}$ can be obtained with a conventional 3D US calibration method~\cite{hsu2009freehand}.

\label{sec:insideOutTracking}
Inside-out tracking is proposed on the foundation of a miniature camera setup as described in Sec.~\ref{sec:experiments}.
The setup provides different image modalities for the visual SLAM. 
Monocular SLAM is not suitable for our needs, since it needs an appropriate translation without rotation within the first frames for proper initialization and suffers from drift due to accumulating errors over time. 
Furthermore, the absolute scale of the reconstructed map and the trajectory is unknown due to the arbitrary baseline induced by the non-deterministic initialization for finding a suitable translation.
The latter is needed to triangulate matched feature points between two views. 
Relying on the depth data from the sensor would not be sufficient for the desired tracking accuracy, due to noisy depth information.
A stereo setup can account for absolute scale by a known fixed baseline. Movements with rotations only can be accounted for with a stereo system, since matched feature points can be triangulated for each frame.

For the evaluations we run experiments with publicly available SLAM methods for better reproducibility and comparability. ORB-SLAM2~\cite{ORB-SLAM2_Mur-Artal_TOR2017} is use as state-of-the-art feature based method. The well-known direct methods~\cite{LSD-SLAM_Engel_ECCV2014,DSO_Engel_PAMI2018} are not eligible due to the restriction to monocular cameras. 
We rely on a the recent publicly available\footnote{https://github.com/JiatianWu/stereo-dso, Horizon Robotics, Inc. Beijing, China, Authors: Wu, Jiatian; Yang, Degang; Yan, Qinrui; Li, Shixin} stereo implementation of Direct Sparse Odometry (DSO)~\cite{wang2017stereoDSO}.

\label{sec:calibration}
The intrinsic camera parameters of the involved monocular and stereo cameras (RGB, IR1, IR2) are estimated as proposed by~\cite{zhang2000flexible}. We use the standard pinhole camera model with two radial distortion coefficients. The stereo geometry is calculated via OpenCV\footnote{https://github.com/itseez/opencv}.
For the rigid transformation from the robotic end effector to the inside-out camera, we use the hand-eye calibration algorithm of Tsai-Lenz~\cite{tsai1989new} in eye-on-hand variant implemented in ViSP~\cite{marchand2005visp} and the eye-on-base version to get the rigid transformation from the optical tracking system to the robot base.
To calibrate the ultrasound image plane with respect to the different tracking systems, we use the open source PLUS ultrasound toolkit~\cite{lasso2014plus} and provide a series of correspondence pairs using a tracked stylus pointer, and retrieve the desired transformation matrix.

\section{Experiments and Validation} 
\label{sec:experiments}
\begin{figure}[t]
\centering
\includegraphics[height=4.cm]{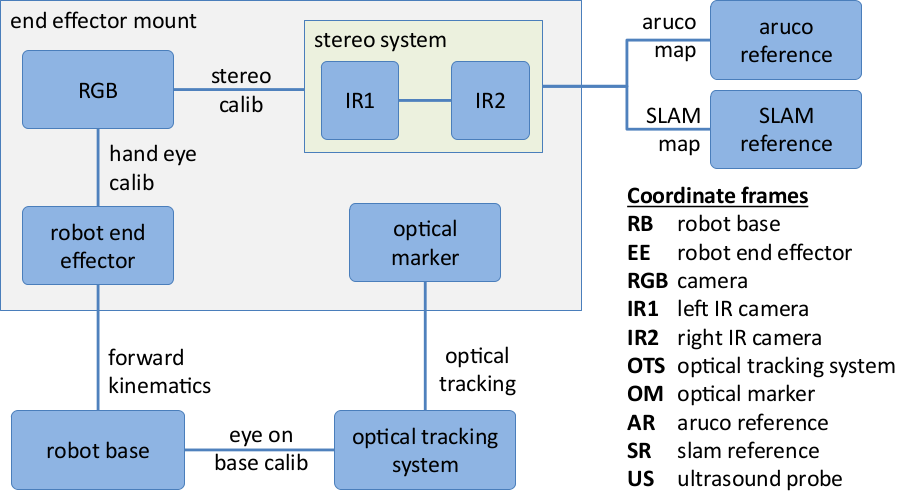}
\includegraphics[height=4.cm]{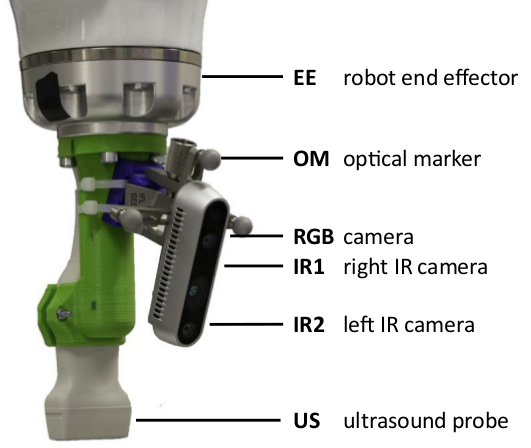}
	\caption{\textbf{System architecture and coordinate frames.} Shown are all involved coordinate reference frames to evaluate the system performance (\textbf{left}) as well as the specific ultrasound mount used for validation, integrating optical and camera-based tracking with one attachable target (\textbf{right}).}
	\label{fig:coordinateFrames}
\end{figure}

\begin{figure}[t]
\centering
\includegraphics[width=\textwidth]{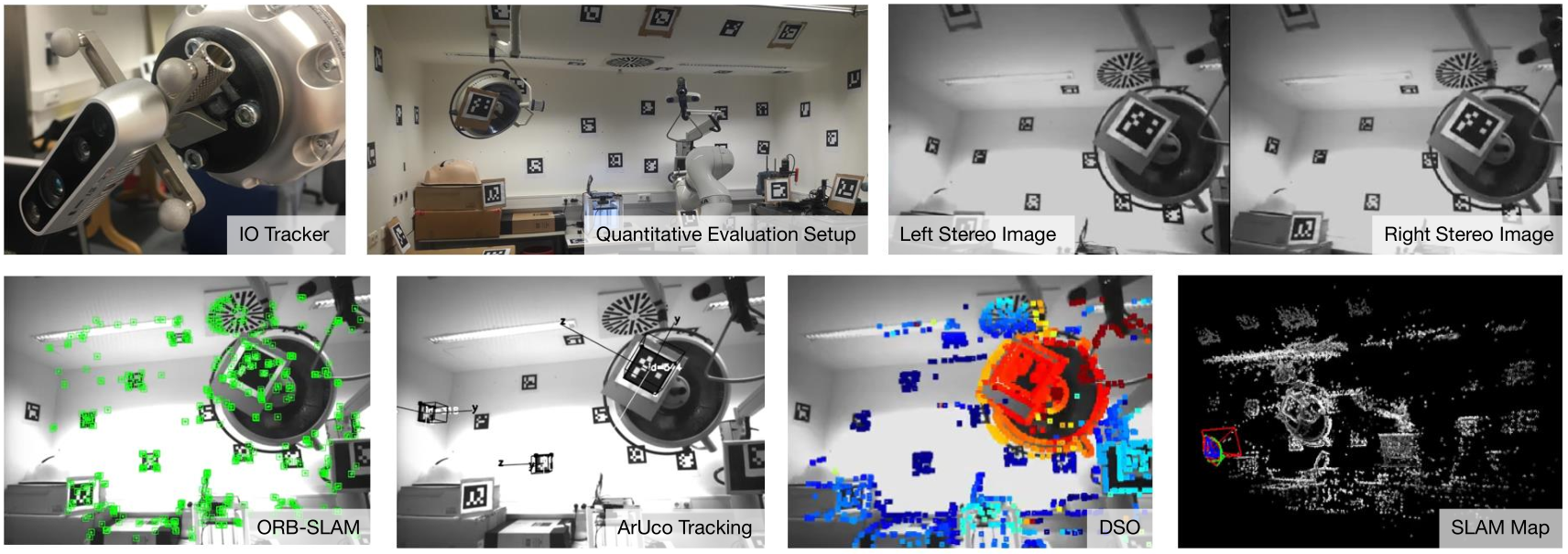}
	\caption{\textbf{Quantitative evaluation setup.} The first row illustrates the operating room where the quantitative analysis is performed together with the inside-out stereo view. The second row depicts various calculated SLAM information necessary to create the map.}
	\label{fig:wetLab}
\end{figure}

To validate the proposed tracking approach, we first evaluate the tracking accuracy, followed by a specific analysis for the suitability to 3D ultrasound imaging.
We use a KUKA iiwa (KUKA Roboter GmbH, Augsburg, Germany) 7 DoF robotic arm to gather ground truth tracking data, as it provides a guaranteed positional reproducibility of $\pm0.1$~mm. 
To provide a realistic evaluation, we also utilize an optical infrared-based outside-in tracking system (Polaris Vicra, Northern Digital Inc., Waterloo, Canada).
Inside-out tracking is performed with the Intel RealSense Depth Camera D435 (Mountain View, US), providing RGB and infrared stereo data in a portable system (see Fig. \ref{fig:coordinateFrames}).
Direct and feature based SLAM methods for markerless inside-out tracking are compared and evaluated against marker based optical inside-out tracking with ArUco markers and classical optical outside-in tracking.
For the former, ArUco~\cite{Aruco2014} markers with a size of $16\times 16$~cm are placed in the acquisition room.
For a quantitative analysis, a combined marker with an optical target and a miniature vision sensor is used (see Fig.~\ref{fig:coordinateFrames}) and attached to the robot end effector.
The robot is controlled using the Robot Operating System (ROS) while the camera acquisition is done on a separate machine using the intel RealSense SDK\footnote{https://github.com/IntelRealSense/librealsense}. 
Images are acquired with a resolution of $640\times 480$ pixels at a frame rate of $30$~Hz.
The pose of the RGB camera and the tracking target are communicated via TCP/IP with a publicly available library\footnote{https://github.com/IFL-CAMP/simple}.
The images are processed on an intel Core i7-6700 CPU, 64bit, 8GB RAM running Ubuntu 14.04.
We use the same constraints as in a conventional TRUS. Thus, the scanning time, covered volume and distance of the tracker is directly comparable and the error analysis reflects this specific procedure with all involved components. Fig.~\ref{fig:wetLab} shows the clinical environment for the quantitative evaluation together with the inside-out view and the extracted image information for the different SLAM methods.

\subsection{Tracking Accuracy}
To evaluate the tracking accuracy without specific application to imaging, we use the setup described above and acquire a series of pose sequences.
The robot is programmed to run in gravity compensation mode such that it can be directly manipulated by a human operator.
The forward kinematics of a robotic manipulator are used as ground truth (GT) for the actual movement.

To allow for error evaluation, we transform all poses of the different tracking systems in the joint coordinate frame coinciding at the RGB-camera of the end effector mount (see Fig. \ref{fig:coordinateFrames} for an overview of all reference frames)
\begin{align}
 \tensor[^{RGB}]{\textbf{T}}{_{RB}}\, &=
 \tensor[^{RGB}]{\textbf{T}}{_{EE}} \cdot
 \tensor[^{EE}]{\textbf{T}}{_{RB}} \\
 \tensor[^{RGB}]{\textbf{T}}{_{SR}}\, &=
 \tensor[^{RGB}]{\textbf{T}}{_{EE}} \cdot \tensor[^{EE}]{\textbf{T}}{_{RB}} \cdot \tensor[^{RB}]{\textbf{T}}{_{IR1,0}} \cdot \tensor[^{IR1,0}]{\textbf{T}}{_{SR}} \\
 \tensor[^{RGB}]{\textbf{T}}{_{AR}}\, &=
 \tensor[^{RGB}]{\textbf{T}}{_{EE}} \cdot \tensor[^{EE}]{\textbf{T}}{_{RB}} \cdot \tensor[^{RB}]{\textbf{T}}{_{IR1,0}} \cdot \tensor[^{IR1,0}]{\textbf{T}}{_{AR}} \\
 \tensor[^{RGB}]{\textbf{T}}{_{OTS}}\, &=
 \tensor[^{RGB}]{\textbf{T}}{_{EE}} \cdot
 \tensor[^{EE}]{\textbf{T}}{_{RB}} \cdot
 \tensor[^{RB}]{\textbf{T}}{_{OTS}} \cdot
 \tensor[^{OTS}]{\textbf{T}}{_{OM}},
\end{align}
providing a direct way to compare the optical tracking system (OTS), to SLAM-based methods (SR), and the ArUco-based tracking (AR).
Multiple calibrations have shown that the residuals from the robotic hand-eye calibration is negligible with respect to the tracking and US calibration.

\begin{figure}[t]
\centering
\includegraphics[width=\textwidth]{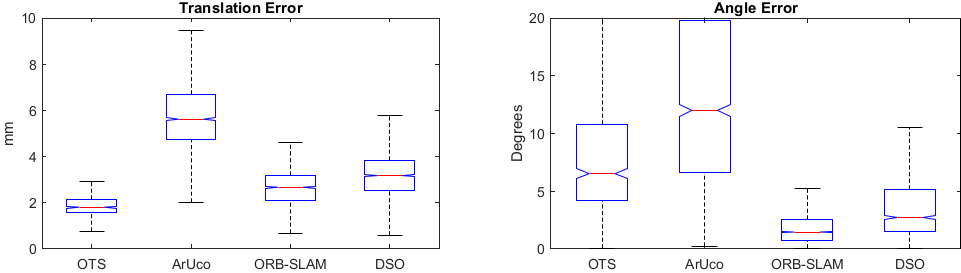}
	\caption{\textbf{Comparison of tracking error.} Shown are translational and rotational errors compared to ground truth for all evaluated systems.}
	\label{fig:evalAcc}
\end{figure}

In overall, 5~sequences were acquired with a total of 8698~poses. 
The pose error for all compared system is indicated in Fig. \ref{fig:evalAcc}, where the translation error is given by the RMS of the residuals compared with the robotic ground truth while the illustrated angle error gives angular deviation of the rotation axis.
From the results it can be observed that compared to GT, optical tracking provides the best results, with translation errors of 1.90~$\pm$~0.53~mm, followed by 2.65~$\pm$~0.74~mm for ORB-SLAM and 3.20~$\pm$~0.96 for DSO, ArUco with 5.73~$\pm$~1.44~mm.
Interestingly, the SLAM-based methods provide better results compared to OTS, with errors of  1.99~$\pm$~1.99$^\circ$ for ORB-SLAM, followed by 3.99~$\pm$~3.99$^\circ$ for DSO, respectively. OTS estimates result in errors of 8.43~$\pm$~6.35$^\circ$, and ArUco orientations are rather noisy with 29.75~$\pm$~48.92$^\circ$.

%

\subsection{Markerless Inside-Out 3D Ultrasound}
\begin{figure}[t]
\centering
\includegraphics[width=\textwidth]{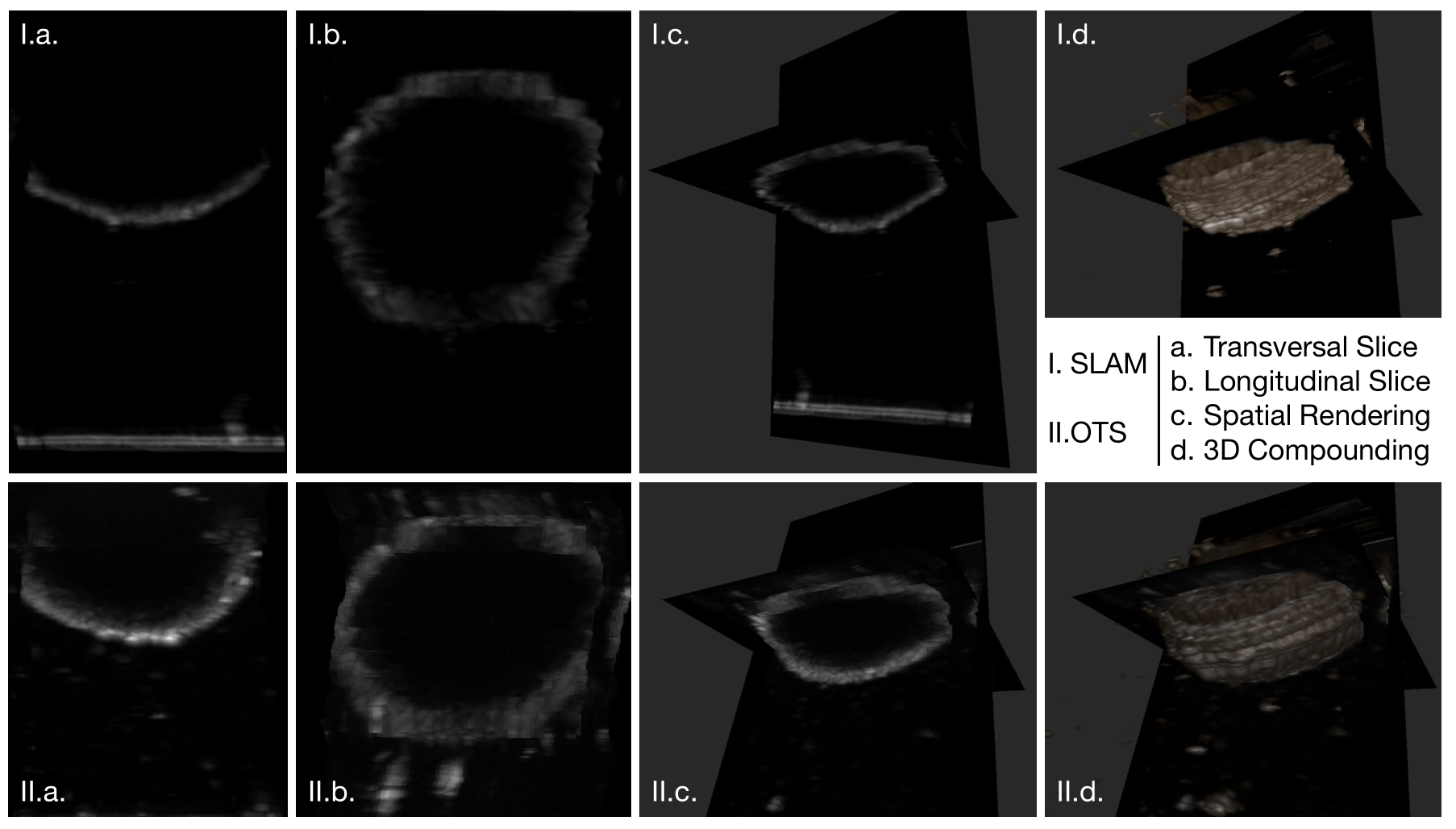}
	\caption{\textbf{Visualization of 3D US compounding quality.} Shown are longitudinal and transversal slices as well as a 3D rendering of the resulting reconstructed 3D data from a tracked ultrasound acquisition of a ball phantom for the proposed tracking using ORB-SLAM in comparison with a commercial outside-in OTS. The structure appears spherically while the rotational accuracy advantage of ORB-SLAM causes a smoother rendering surface and a more clearly defined phantom boundary in the computed slices.}
	\label{fig:evalUS_Comp}
\end{figure}

On the foundation of favourable tracking characteristics, we evaluate the performance of a markerless inside-out 3D ultrasound system by means of image quality and reconstruction accuracy for a 3D US compounding.
For imaging, the tracking mount shown in Fig. \ref{fig:coordinateFrames} is integrated with a 128 elements linear transducer (CPLA12875, 7~MHz) connected to a cQuest Cicada scanner (Cephasonics, CA, USA).
For data acquisition, a publicly available real-time framework is employed\footnote{https://github.com/IFL-CAMP/supra} in conjunction with ROS, and calibration is performed using a stylus as described in Sec. \ref{sec:calibration}.
We perform a sweep acquisition, comparing OTS outside-in tracking with the proposed inside-out tracking and evaluate the quality of the reconstructed data while we deploy~\cite{Busam2016} for temporal pose synchronization.
Fig.~\ref{fig:evalUS_Comp} shows a qualitative comparison of the 3D US compoundings for the same sweep with the different tracking methods.

\section{Discussion and Conclusion}
\label{sec:discussion}

From our evaluation, it appears that Aruco markers are viable only for approximate positioning within a room rather than accurate tracking.
Our proposed inside-out approach shows valuable results compared to standard OTS and even outperforms the outside-in system in terms of rotational accuracy. 
These findings concur with assumptions based on the camera system design, as small rotations close to the optical principal point of the camera around any axis will lead to severe changes in the viewing angle, which can visually be described as inside-out rotation leverage effect.

One main advantage of the proposed methods is with respect to usability in practice. 
By not relying on specific markers, there is no need for setting up an external system or a change in setup during procedures. 
Additionally, we can avoid line-of-sight problems, and potentially allow for highly accurate tracking even for complete rotations around the camera axis without loosing tracking.
Besides the results above, our proposed method is capable of orientating itself within an unknown environment by mapping its surrounding from the beginning of the procedure. This mapping is build up from scratch without the necessity of any additional calibration. Our tracking results for a single sensor also suggest further investigation towards collaborative inside-out tracking with multiple systems at the same time, orientating themselves within a global map as common reference frame.

Overall, we presented a markerless inside-out tracking method based on visual SLAM and demonstrated its accuracy for general tracking as well as 3D ultrasound imaging.
Reasoned by the accuracy and versatility, we hope that this will lead to a more detailed investigation of the proposed markerless inside-out tracking method for the use in medical procedures also by other research groups.
This is in particular interesting for applications that include primarily rotation such as transrectal prostate fusion biopsy.

\bibliographystyle{splncs}
\bibliography{literature}

\begin{thebibliography}{10}

\bibitem{hennersperger2017towards}
Hennersperger, C., Fuerst, B., Virga, S., Zettinig, O., Frisch, B., Neff, T.,
  Navab, N.:
\newblock Towards {MRI}s-based autonomous robotic {US} acquisitions: a first
  feasibility study.
\newblock IEEE transactions on medical imaging \textbf{36}(2) (2017)  538--548

\bibitem{Adebar2011}
Adebar, T., Salcudean, S., Mahdavi, S., Moradi, M., Nguan, C., Goldenberg, L.:
\newblock A robotic system for intra-operative trans-rectal ultrasound and
  ultrasound elastography in radical prostatectomy.
\newblock In: International Conference on Information Processing in
  Computer-Assisted Interventions, Springer (2011)  79--89

\bibitem{kral2013comparison}
Kral, F., Puschban, E.J., Riechelmann, H., Freysinger, W.:
\newblock Comparison of optical and electromagnetic tracking for navigated
  lateral skull base surgery.
\newblock The International Journal of Medical Robotics and Computer Assisted
  Surgery \textbf{9}(2) (2013)  247--252

\bibitem{busam2015stereo}
Busam, B., Esposito, M., Che'Rose, S., Navab, N., Frisch, B.:
\newblock A stereo vision approach for cooperative robotic movement therapy.
\newblock In: Proceedings of the IEEE International Conference on Computer
  Vision Workshops. (2015)  127--135

\bibitem{heuveling2012sentinel}
Heuveling, D., Karagozoglu, K., Van~Schie, A., Van~Weert, S., Van~Lingen, A.,
  De~Bree, R.:
\newblock Sentinel node biopsy using 3d lymphatic mapping by freehand spect in
  early stage oral cancer: a new technique.
\newblock Clinical Otolaryngology \textbf{37}(1) (2012)  89--90

\bibitem{fenster2001three}
Fenster, A., Downey, D.B., Cardinal, H.N.:
\newblock Three-dimensional ultrasound imaging.
\newblock Physics in medicine \& biology \textbf{46}(5) (2001)  R67

\bibitem{esposito2015cooperative}
Esposito, M., Busam, B., Hennersperger, C., Rackerseder, J., Lu, A., Navab, N.,
  Frisch, B.:
\newblock Cooperative robotic gamma imaging: Enhancing us-guided needle biopsy.
\newblock In: International Conference on Medical Image Computing and
  Computer-Assisted Intervention, Springer (2015)  611--618

\bibitem{sun2014probe}
Sun, S.Y., Gilbertson, M., Anthony, B.W.:
\newblock Probe localization for freehand 3d ultrasound by tracking skin
  features.
\newblock In: International Conference on Medical Image Computing and
  Computer-Assisted Intervention, Springer (2014)  365--372

\bibitem{ORB-SLAM2_Mur-Artal_TOR2017}
Mur-Artal, R., Tardós, J.D.:
\newblock {ORB-SLAM2}: An open-source slam system for monocular, stereo, and
  {RGB-D} cameras.
\newblock IEEE Transactions on Robotics \textbf{33}(5) (2017)  1255--1262

\bibitem{LSD-SLAM_Engel_ECCV2014}
Engel, J., Sch\"ops, T., Cremers, D.:
\newblock {LSD-SLAM}: Large-scale direct monocular {SLAM}.
\newblock In: European Conference on Computer Vision. (2014)

\bibitem{DSO_Engel_PAMI2018}
Engel, J., Koltun, V., Cremers, D.:
\newblock Direct sparse odometry.
\newblock Transactions on Pattern Analysis and Machine Intelligence (2018)

\bibitem{hsu2009freehand}
Hsu, P.W., Prager, R.W., Gee, A.H., Treece, G.M.:
\newblock Freehand 3d ultrasound calibration: a review.
\newblock In: Advanced imaging in biology and medicine.
\newblock Springer (2009)  47--84

\bibitem{wang2017stereoDSO}
Wang, R., Schw\"orer, M., Cremers, D.:
\newblock Stereo {DSO}: Large-scale direct sparse visual odometry with stereo
  cameras.
\newblock In: International Conference on Computer Vision. (2017)

\bibitem{zhang2000flexible}
Zhang, Z.:
\newblock A flexible new technique for camera calibration.
\newblock IEEE Transactions on pattern analysis and machine intelligence
  \textbf{22}(11) (2000)  1330--1334

\bibitem{tsai1989new}
Tsai, R.Y., Lenz, R.K.:
\newblock A new technique for fully autonomous and efficient 3d robotics
  hand/eye calibration.
\newblock IEEE Transactions on robotics and automation \textbf{5}(3) (1989)
  345--358

\bibitem{marchand2005visp}
Marchand, {\'E}., Spindler, F., Chaumette, F.:
\newblock Visp for visual servoing: a generic software platform with a wide
  class of robot control skills.
\newblock IEEE Robotics \& Automation Magazine \textbf{12}(4) (2005)  40--52

\bibitem{lasso2014plus}
Lasso, A., Heffter, T., Rankin, A., Pinter, C., Ungi, T., Fichtinger, G.:
\newblock Plus: open-source toolkit for ultrasound-guided intervention systems.
\newblock IEEE Transactions on Biomedical Engineering \textbf{61}(10) (2014)
  2527--2537

\bibitem{Aruco2014}
Garrido-Jurado, S., noz Salinas, R.M., Madrid-Cuevas, F., Mar\'in-Jim\'enez,
  M.:
\newblock Automatic generation and detection of highly reliable fiducial
  markers under occlusion.
\newblock Pattern Recognition \textbf{47}(6) (2014)  2280 -- 2292

\bibitem{Busam2016}
Busam, B., Esposito, M., Frisch, B., Navab, N.:
\newblock Quaternionic upsampling: Hyperspherical techniques for 6 dof pose
  tracking.
\newblock In: 3D Vision (3DV), 2016 Fourth International Conference on, IEEE
  (2016)  629--638

\end{thebibliography}

\end{document}